\documentclass{article}
\usepackage{spconf,amsmath,graphicx}
\usepackage{tabularx,booktabs,amssymb}
\usepackage[dvipsnames]{xcolor}
\usepackage{hyperref}

\definecolor{new}{HTML}{6C8EBF}
\usepackage[T1]{fontenc} 

\title{Self-supervised Learning on Camera Trap Footage \\yields a strong universal face embedder}

\name{Vladimir Iashin $\quad$ Horace Lee $\quad$ Dan Schofield$\quad$ Andrew Zisserman\thanks{This research was funded by EPSRC Programme Grant VisualAI EP/ T028572/1. We thank the Guinean authorities (DGERSIT \& IREB), T. Matsuzawa, Kyoto U., and contributors to the Bossou dataset, Tacugama Chimpanzee Sanctuary, local authorities, and field staff for access to Loma Mts camera trap data and support with data sharing and conservation.}}
\address{Visual Geometry Group (VGG), Department of Engineering Science, University of Oxford}

\begin{document}
\maketitle
\begin{abstract}
Camera traps are revolutionising wildlife monitoring by capturing vast amounts of visual data; however, the manual identification of individual animals remains a significant bottleneck. This study introduces a fully self-supervised approach to learning robust chimpanzee face embeddings from unlabeled camera-trap footage. Leveraging the DINOv2 framework, we train Vision Transformers on automatically mined face crops, eliminating the need for identity labels. Our method demonstrates strong open-set re-identification performance, surpassing supervised baselines on challenging benchmarks such as Bossou, despite utilising no labelled data during training. This work underscores the potential of self-supervised learning in biodiversity monitoring and paves the way for scalable, non-invasive population studies. \\
\makebox[\columnwidth][c]{\href{https://www.robots.ox.ac.uk/~vgg/research/ChimpUFE/}{\color{new} \textbf{\texttt{robots.ox.ac.uk/\textasciitilde vgg/research/ChimpUFE}}}
}

\end{abstract}
\section{Introduction}
\label{sec:intro}

Autonomous camera-trap networks are rapidly becoming the work-horse of terrestrial ecological monitoring \cite{debetencourt2024camera}. 
Unlike labour-intensive techniques such as genetic sampling, transect surveys, or long-term habituation, camera traps are inexpensive, unobtrusive, minimally invasive, scalable, and able to record for months in habitats otherwise hard to observe. 
Their images and videos already underpin long-term studies and monitoring of demography, behaviour, and disease in great apes, yet the crucial step of \textit{individual identification} still relies on specialists manually scanning thousands of frames~\cite{debetencourt2024camera}.  
With populations often exceeding a thousand individuals in a single geographical area, this manual workflow is the main bottleneck for estimating community size and connectivity at landscape scale.  
Recent proof-of-concept systems for primate re-identification demonstrate that deep learning can remove this bottleneck \cite{guo2020automatic}, but they inherit two major limitations from the human face-recognition models they adapt.

First, state-of-the-art human face recognisers such as CosFace \cite{wang2018cosface} and ArcFace \cite{deng2019arcface} achieve near-perfect accuracy by training on hundreds of images per identity and hundreds of thousands of identities, e.g.\ MS-Celeb-1M \cite{guo2016ms}. 
For non-human species, the largest curated face datasets (e.g.\ C-Tai, C-Zoo) contain only a few dozen individuals \cite{freytag2016chimpanzee}, and each new label requires expert annotation.  
As a result, prior chimpanzee models operate only in captivity or on small wild sub-populations, limiting ecological utility.

Second, most off-the-shelf face-recognition models are trained for the \textit{closed-set} scenario, where every probe image is guaranteed to depict an identity present in the training data. 
Camera-trap monitoring violates this premise: recordings inevitably contain chimpanzees that have never been annotated before. 
A practical model must therefore handle verification and re-identification of known individuals and remain robust when confronted with entirely new ones, a setting usually referred to as \textit{open-set} recognition \cite{grother2003face,scheirer2012toward}.

In parallel, computer vision research has witnessed a leap in representation learning.  
Self-supervised Vision Transformers trained with DINO \cite{caron2021emerging} and its successor DINOv2 \cite{oquab2023dinov2} learn feature spaces that support retrieval, clustering, and few-shot transfer without any manual labels.  
Crucially, the metric structure from these objectives is well suited to open-set verification and re-identification, where test-time identities may be absent during pre-training.  
These properties match the constraints of wildlife monitoring, yet remain unexplored for chimpanzee identification at scale.

Our objective is to obtain an encoder that can be used for open-set identification, by using self-supervised learning at scale. We envisage that it will perform open-set identification using a straightforward $k$-nearest-neighbour search in the embedding space: each probe face is mapped to its $k$ closest labelled exemplars, and the resulting weighted neighbourhood vote assigns the identity. 
This approach scales seamlessly from few-shot labelling to population-level surveys.

\begin{figure*}[t]
    \centering
    \includegraphics[width=\linewidth]{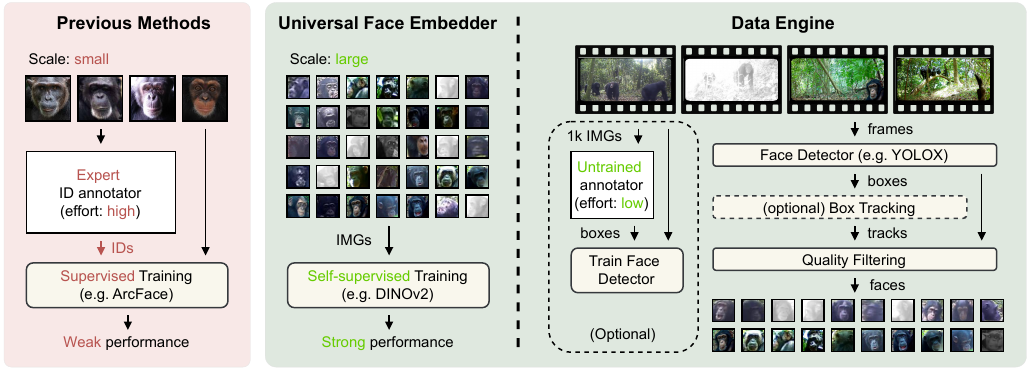} 
    \caption{ 
    \textbf{Overview of our proposed pipeline for learning universal chimpanzee face embeddings}.
    \textit{Left}: Previous methods rely on high-effort expert ID annotation and supervised training (e.g.\ ArcFace), resulting in limited scalability and weaker performance in open-set conditions.
    \textit{Middle}: Our Universal Face Embedder leverages large-scale, unlabelled training data and a self-supervised model (e.g. DINOv2) to learn a retrieval-friendly embedding with strong performance.
    \textit{Right}: The supporting Data Engine extracts high-quality face crops from raw camera-trap videos using a lightweight detector, optional tracker, and low-effort annotation which eliminates the need for identity labels.}
    \label{fig:approach}
\end{figure*}

With this objective in mind, we contribute the first \textit{universal, open-set chimpanzee face embedder}--one that performs robustly across diverse chimpanzee scenarios, including both captive and wild environments, and remains agnostic to variations in image quality--trained \textit{entirely} with self-supervision on unlabelled camera-trap footage.
Concretely, first, we assemble the largest in-the-wild chimpanzee face corpus to date by automatically mining $\sim$0.6 M high-quality face crops from $\sim$90 hours of trap videos. 
Secondly, to address different computational constraints, we pre-train two variants of ViT backbone (22 M and 87 M) with the DINOv2 objective on this corpus, that run at 100+ fps on a single NVIDIA A4000, 4$\times$ faster than the current state of the art face embedder.

\section{Related Work}
\label{sec:related_work}

Early chimpanzee ID studies adapted human face pipelines to small, labelled datasets: Loos \& Ernst \cite{loos2013automated} fused global–local handcrafted descriptors on the captive ChimpZoo and semi-wild ChimpTa\"{i} sets, but their method required meticulous alignment and ID supervision.  
Freytag et al.\ \cite{freytag2016chimpanzee} introduced Log-Euclidean CNNs that jointly predicted identity, age, and sex on the same benchmarks, establishing the first deep-learning baseline for primates.  
Deb et al.'s PrimNet \cite{deb2018face} later added open-set re-ID for three primate species, and Schofield et al.\ \cite{schofield2019chimpanzee} scaled CNN-based recognition to 14 years of Bossou footage by coupling detection and tracking, though still with exhaustive manual labels.

After ArcFace transformed human face recognition, wildlife research followed suit.  
PetFace benchmarks 257k individuals from 13 taxa with ResNet \cite{he2016deep} backbones \cite{shinoda2024petface};  
MegaDescriptor pairs ArcFace with a Swin encoder \cite{liu2021swin} trained on 29 wildlife datasets \cite{vcermak2024wildlifedatasets};  
and MiewID-msv3 brings ArcFace to an EfficientNetV2 \cite{liu2021swin} trained on 64-species data from the Wildbook platform \cite{WildMe2023}.

All of these approaches remain fully supervised.  
Even the most recent attempt to exploit camera-trap footage, a ViT fine-tuned with triplet loss on a few dozen \emph{gorilla} IDs by Laskowski et al.\ \cite{Laskowski2023GorillaVision}, relies on annotated identities to form positive pairs, binding performance to a small fixed vocabulary.  
In contrast, our universal chimpanzee embedder eliminates identity labels during training altogether: it learns directly from \emph{unlabelled} in-the-wild face tracks, enabling open-set generalisation to the thousands of individuals encountered in landscape-scale camera-trap surveys.

\section{Universal Chimpanzee Face Embedder}
\label{sec:method}

This section describes how we scale face representation learning beyond the limits of conventional identity-supervised pipelines. 
The approach is outlined in Figure~\ref{fig:approach}.
 §\ref{sec:model} Model Design shows how a self-supervised backbone such as DINOv2 is adapted to mined faces to yield a universal, open-set chimpanzee embedding that requires no individual ID labels at training time.
§\ref{sec:dataset} Data Engine will then explain how we mine hundreds of thousands of chimpanzee face crops directly from raw camera-trap videos with only light annotation effort.

\subsection{Model Design}
\label{sec:model}

\noindent\textbf{Problem formulation.} 
During inference, given a probe face image of an \emph{unknown} chimpanzee, we wish to retrieve all instances of the same individual that exist in a gallery of previously collected (and annotated) images.  
This \emph{open-set re-identification} (re-ID) setting differs from \textit{closed-set} face recognition because 
(i) gallery membership is not fixed as new individuals may appear at any time, 
and (ii) we do not assume class‐label supervision during training.  
Instead, we learn a function $f_\theta$, mapping a face image ($\mathbb{R}^{H\times W\times 3}$) to a $d$-dimensional representation, which is used to compute cosine similarity between faces at test time, allowing verification and reducing re-ID to $k$-NN search in that space.

\noindent\textbf{Design requirements.}  
We require our embedding to satisfy three core requirements: 
(1) The training approach should leverage large-scale camera trap footage, that has no labels provided; 
(2) The model should scale gracefully with larger training datasets and inference hardware; 
(3) The learned representation should exhibit retrieval-friendly geometry so that cosine similarity yields faithful neighbourhoods and makes index structures, e.g.\ FAISS \cite{douze2024faiss}, effective.

\noindent\textbf{Candidate embedding learning approaches.}  
Supervised margin-based objectives such as CosFace \cite{wang2018cosface} and ArcFace \cite{deng2019arcface} already satisfy conditions (2) and (3) by explicitly maximizing inter-class margins and minimizing intra-class variance, though they rely on labelled data and hence lie outside the fully self-supervised scope we target. 
Contrastive methods such as SimCLR \cite{chen2020simple} and MoCo v3 \cite{chen2021empirical} satisfy conditions (1) and (2), but they optimise primarily for inter-image discrimination and still rely on very large batches or a queue-based memory bank in earlier MoCo variants to supply negatives and prevent representation collapse. 
Masked-image modelling approaches like MAE \cite{he2022masked} and iBOT \cite{zhou2021ibot} excel at pixel-level reconstruction, yet their encoders need an extra decoder/projection head during pre-training and typically require task-specific fine-tuning before matching contrastive models on $k$-NN retrieval. 

Recently, self-distillation frameworks such as DINO \cite{caron2021emerging} and its successor DINOv2 \cite{oquab2023dinov2} have emerged as especially promising for our goals. 
These models train a \textit{student} network to match the softmax outputs of a momentum-updated teacher across \textit{global} and \textit{local} image crops, using techniques like \emph{centring} (running-mean subtraction) and \emph{sharpening} (low-temperature scaling) of teacher logits to prevent collapse. 
DINOv2 further introduces a patch-level alignment loss and a uniformity regulariser, both of which help produce embeddings that are highly amenable to retrieval tasks, such as open-set identification with $k$-NN search. 
Notably, these methods achieve strong zero-shot and retrieval performance without any manual labels during pre-training, making them well suited for wildlife monitoring at scale and particularly effective for the large volumes of unlabelled camera-trap footage available in our setting.

\begin{table}[t]
  \centering
  \footnotesize
  \setlength{\tabcolsep}{3pt}  
  \begin{tabular}{lrrrrr}
    \toprule
    \textbf{Source} & \textbf{\#Videos} & \textbf{Frames} & \textbf{Resolution} & \textbf{Raw dets.} & \textbf{Filtered dets.} \\
    \midrule
    PanAf-20K   & 20473 & 7.0 M & 404 p & 3.0 M &  0.3M \\
    Loma Mts.     & 491 & 0.8 M & 1080 p & 0.3 M    &  0.3M \\
    \bottomrule
  \end{tabular}
  \caption{Camera-trap corpora mined by our data engine.  \textbf{Raw dets.}~= face boxes before filtering.}
  \label{tab:corpora}
  \vspace{-0.8em}
\end{table}

\subsection{Data Engine}
\label{sec:dataset}
Camera-trap footage is plentiful, cheap to collect and records wild animals in their natural habitat by day and night, yet it remains under-used because traditional ID annotation scales poorly: a curator must already recognise every individual in the archive, or rely on name–image pairs scraped from the internet as MS-Celeb-1M does for humans, which is unavailable for animals in the wild. 
We instead \textit{mine in-the-wild footage for faces} and use them to train a strong self-supervised model.
Our mining pipeline, as illustrated in Figure~\ref{fig:approach} (right), has two modular components: Face Detector and Bounding-box Tracker (optional but beneficial).

\noindent\textbf{Face Detector.}
Any off-the-shelf open-source detector works \cite{kuzdeuov2023anyface}, or one can train a lightweight (e.g.\ YOLOX \cite{ge2021yolox}), task-specific model by annotating $\sim$1000 frames without primatology expertise. 
We found that a dedicated detector trained on $\sim$2000 manually-boxed frames captured uncommon viewpoints (high yaw/pitch angles, backs of heads) better than generic models such as AnyFace  \cite{kuzdeuov2023anyface}, which in turn helps the tracker to reconnect trajectories after short occlusions or when a chimp briefly turns its head away from the camera.
Back-of-head crops generally receive low confidence scores and can be filtered out downstream without harming~recall.

\noindent\textbf{Bounding-box Tracker.}
We run ByteTrack \cite{zhang2022bytetrack}, which associates all detections, including low-score ones, across frames, recovering occluded faces, lengthening trajectories, suppressing isolated false positives, and providing track-level identity hints that may supplement contrastive objectives with relations between elements in a batch.

\noindent\textbf{Mined Chimpanzee Face Corpora.}
We apply the Data Engine to two contrasting camera-trap collections (see Table~\ref{tab:corpora} for quantitative details). 
\textit{PanAf-20K} \cite{brookes2024panaf20k} spans 18 Central-African sites; clips are 15s long, relatively low-res at 404 p ($>\!90\,\%$ of footage), and occasionally contain gorillas.  
\textit{Loma Mountains National Park}, Sierra Leone, \cite{molina2023reaffirming} uses 1080 p traps that record 60s by day and 20s by night.  
Because motion triggers only at the clip start, raw detections are far fewer than frames, even when multiple chimpanzees appear.

\begin{figure}[t]
    \centering
    \includegraphics[width=\linewidth]{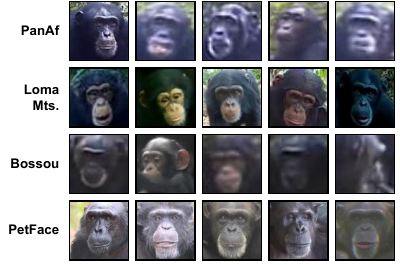} 
    \caption{Example images from \emph{PanAf}, \emph{Loma Mts.} (training), and \emph{Bossou}, \emph{PetFace} (evaluation). Wild footage in \emph{PanAf}, \emph{Loma Mts.}, and \emph{Bossou} has unaligned faces, extreme poses, and variable resolution due to fluctuating camera-subject distance. In contrast, \emph{PetFace} offers close-up, well-aligned images under more controlled conditions. 
    }
    \label{fig:samples}
\end{figure}

After detection and tracking, we sorted the resulting face detections by confidence and observed a large difference in the proportion of usable faces between datasets: high-quality face crops exceed 90\,\% in Loma but reach only $\sim$10\,\% in PanAf. 
We attribute this gap to lower resolution in PanAf footage rather than a limitation of the detector itself.
To balance quality without collapsing track diversity, we keep the most-confident 20\,\% of PanAf detections, then randomly subsample half of that set, yielding $\sim$310k faces, about 10\,\% of the original detections, whose quality matches Loma's (albeit at lower resolution). See examples in Figure~\ref{fig:samples}.

A modest detector and a generic tracker can transform large, heterogeneous trap archives into a diverse, in-the-wild dataset for self-supervised training.  
The accompanying metadata (time, location, track continuity) could further provide positive/negative pairs, though our non-contrastive method does not yet rely on it.

\begin{table}[t]
\footnotesize
\centering
\begin{minipage}{0.32\columnwidth}
\centering
\setlength{\tabcolsep}{2pt}
\begin{tabular}{lr}
\toprule
\textbf{Data} & \textbf{Train} \\
\midrule
PanAf-20K (\textbf{P}) & 314k\\
Loma Mts.\ (\textbf{L}) & 314k\\
Bossou-14 (\textbf{B}) & 335k\\
\bottomrule
\end{tabular}
\end{minipage}%
\hfill
\begin{minipage}{0.64\columnwidth}
\centering
\setlength{\tabcolsep}{3pt}
\begin{tabular}{lr|cc|c}
\toprule
 &  &
\multicolumn{2}{c|}{\textbf{Re-ID}} & 
\multicolumn{1}{c}{\textbf{Verif.}} \\
\textbf{Data}& \textbf{IDs}& \textbf{Gallery} & \textbf{Queries} & \textbf{+/--} \\
\midrule
Bossou-9        & 9   & 3150 & 630  & 7749 \\
PetFaceC$\star$ & 376 & 2477 & 376  & 15205 \\
\bottomrule
\end{tabular}
\end{minipage}
\caption{Datasets used for training (left) and evaluation (right), with statistics for re-ID and verification.}
\label{tab:datasets}
\end{table}

\section{Datasets and Evaluation}
\label{sec:datasets}

\noindent\textbf{Training mixture.}
We mine two balanced corpora with the Data Engine in §\ref{sec:dataset}:  
\textit{PanAf-20K} and \textit{Loma Mountains} each contribute $\sim$314k high-quality face crops. 
Beyond PanAf and Loma, we also use $\sim$335k face crops from the annotated \textit{Bossou} video archive~\cite{schofield2019chimpanzee}, containing 23 identities.  
Although Bossou footage was captured with camcorders on tripods by humans rather than motion-triggered traps, it shows wild chimpanzees, making it a valuable complement for both training and evaluation.  
To guarantee a clean split, we first remove every track that belongs to the nine individuals reserved for evaluation, leaving the remaining 14 IDs.  
All identity tags are then discarded and this unlabeled subset is used solely for pre-training; we refer to it as \textit{Bossou-14}.
In total, our training mixture contains $\sim$964k unlabeled images.
The datasets are summarised in Table~\ref{tab:datasets} (left) and face samples from these datasets are shown in Figure~\ref{fig:samples}.

\noindent\textbf{Evaluation benchmarks.}
We evaluate on two benchmarks, Bossou-9 and PetFaceC$\star$ (376 IDs), summarised in Table~\ref{tab:datasets}~(right).  
Bossou~\cite{schofield2019chimpanzee} provides in-the-wild, \emph{unaligned} chimpanzee face crops with substantial pose and quality variation, making it a challenging testbed. 

\noindent\textbf{Re-ID protocol.}
\textit{Open-set re-identification (Re-ID)} asks whether a query face appears in a gallery, crucial when individuals are unseen at training time. 
To this end, for each of 9 IDs in Bossou we randomly sample 10 frames from 35 tracks (3150 images) to build the $k$-NN gallery and keep the remaining 7 tracks (630 images) as queries, yielding \textit{Bossou-9}.
For PetFaceC$\star$ we remove near-duplicates, discard faulty images ($\sim$12\,\%) and classes with fewer than 4 usable portraits from PetFace-Chimp, leaving 2853 images across 376 IDs; for each ID, all but one image form the gallery and the held-out image becomes the query.
Each query is embedded and cosine-matched to gallery embeddings.
We report class-averaged $k$-NN accuracy, sweeping \(k\!\in\!\{\,1,3,5,7,10,20,50\,\}\) to explore the bias-variance trade-off, selecting the best $k$ on a held-out split as final evaluation.
Results are averaged over 10 random splits.

\noindent\textbf{Verification protocol.}
\textit{Open-set verification} judges if two faces depict the same chimp without assuming either image belongs to a known identity.
To this end, for Bossou we equalise track counts by subsampling every individual to the minimum of 42 tracks, select one frame per track (378 images), and generate a balanced set of positive/negative pairs, producing the 7749 pairs reported in Table~\ref{tab:datasets}.  
For PetFaceC$\star$ the same pairing strategy applied to the 2853 portraits yields the 15205 pairs listed in the table.  
We evaluate with ROC-AUC computed from cosine-similarity scores and average the results over 10 random negative sets.

\noindent\textbf{Implementation Details.}
We initialise the backbone from scratch. We found that the general-purpose pre-trained DINOv2 weights did not noticeably improve performance for this task, and follow the ``\texttt{\_short}'' DINOv2 recipe: 
AdamW, base-LR $4\times10^{-3}$, cosine schedule with warmup, weight-decay 0.04, $2 \times 224$ px \emph{global} and $ 8 \times 96 $ px \emph{local} crops, and teacher-momentum ramp-up.
ViT-Small (22 M parameters) is trained for 250k iterations with 64 faces per batch on a single RTX A4000 (16 GB) for 36 hours.
ViT-Base (87 M parameters) uses 14 px patches instead of 16 px, and trains for 500k iterations on one RTX A6000 (48 GB) for $\sim$4 days.
Unlike DINOv2, we perform no heavy data curation for training: every detected face crop (blurred, occluded, or low-light) is kept, except for the basic confidence filter described in §\ref{sec:dataset}

\section{Experiments}
\label{sec:experiments}

\subsection{Baselines}
\label{sec:baselines}
Because no prior chimpanzee embedder is self-supervised, we compare against the strongest \emph{supervised} animal face descriptors: 
(i) \textit{PetFace} (ArcFace ResNet50) \cite{shinoda2024petface} trained on the chimpanzee subset of PetFace;
(ii) \textit{MegaDescriptor} (ArcFace Swin-L) \cite{vcermak2024wildlifedatasets} trained on 29 wildlife multi-species datasets; 
(iii) \textit{MiewID-msv3} (ArcFace EfficientNetV2-M) \cite{WildMe2023} trained on a mixture of datasets covering 64 species.

\begin{table}[t]
    \centering
    \footnotesize
    \setlength{\tabcolsep}{1pt}
    \begin{tabular}{lcrrrrrr}
        \toprule
               ~         &      ~        &        ~        &   \multicolumn{2}{c}{\textbf{Re-identification}} & \multicolumn{2}{c}{\textbf{Verification}} \\
        \textbf{Method } & \textbf{Res.} & \textbf{Params} & \textbf{Bossou-9} & \textbf{PetFaceC$\star$} & \textbf{Bossou-9} & \textbf{PetFaceC$\star$} \\
        \midrule
        \multicolumn{7}{l}{\textit{Supervised baselines}} \\
        PetFace \cite{shinoda2024petface}      & 224         & 24 M  & 49.6 & --\;\(^\dagger\) & 57.5 & --\;\(^\dagger\) \\
        MegaDescr. \cite{vcermak2024wildlifedatasets}  & 384 &229 M & 51.1 & 26.2   & 64.5 & 71.7 \\
        MiewID-ms \cite{WildMe2023}                  & 440 & 51 M  & 56.7 & \textbf{49.4}   & 61.5 & 74.6 \\
        \midrule
        \multicolumn{7}{l}{\textit{Self-supervised, SimCLR}} \\
        Ours (\textbf{L+P+B})                         & 224 & 22 M & 62.1 & 16.0  & 65.5 & 62.5 \\
        \midrule
        \multicolumn{7}{l}{\textit{Self-supervised, DINOv2}} \\
        Ours (\textbf{L})                             & 224 & 22 M & 69.4 & 36.6   & 64.4 & 72.7 \\
        Ours (\textbf{P})                             & 224 & 22 M & 73.5 & 34.2   & 67.3 & 70.1 \\
        Ours (\textbf{L+P})                           & 224 & 22 M & 74.6 & 35.7   & 67.6 & 71.3 \\
        Ours (\textbf{L+P+B})                         & 224 & 22 M & 78.1 & 39.3   & 71.8 & 73.7 \\
        Ours (\textbf{L+P+B})            & 224 & 87 M & \textbf{82.2} & 45.9   & \textbf{74.2} & \textbf{76.3} \\
        \bottomrule
    \end{tabular}
    \vspace{-3ex}
\caption{\textbf{Open-set re-identification (Re-ID) and verification.}  
Class-averaged $k$-NN accuracy (Re-ID) and ROC-AUC (verification) on Bossou-9 and on the cleaned, chimp-only PetFaceC$\star$ split.  
Our self-supervised DINOv2 models—trained \emph{without} identity labels on Loma (\textbf{L}), PanAf (\textbf{P}), and Bossou-14 (\textbf{B}) beat all supervised baselines in three of the four settings, establishing new SOTA on Bossou-9 Re-ID and PetFaceC$\star$ verification.  
\textbf{Res.} = input size; \textbf{Params} = model parameters.  
PetFace-ArcFace (\(^\dagger\)) is omitted on PetFaceC$\star$ because its public checkpoint was partially trained on these images, causing split contamination.}
    \label{tab:main_results}
    \vspace{-3ex}
\end{table}

\subsection{Main Results}
\vspace{-1ex}
\label{sec:main_results}

Table~\ref{tab:main_results} summarises our results.  
Using DINOv2 with mixed wild footage (\textit{L+P+B}, 22 M parameters) we obtain 78.1\,\% Re-ID on Bossou-9, already \textit{21.4 pp} above the best supervised model (MiewID-msv3, 56.7\,\%).  
Scaling to 87 M parameters pushes Bossou-9 Re-ID to 81.6\,\% and lifts verification to 74.9\,\%, again exceeding the supervised baseline.

On PetFaceC$\star$, our best DINOv2 model achieves 43.9\,\% Re-ID and 76.4\,\% verification. 
While our self-supervised approach does not yet match the supervised baseline in re-identification on this portrait-style benchmark, it is trained entirely without identity labels. 
We expect that scaling up the diversity and volume of wild footage will further improve results on such benchmarks. 
Crucially, our model is tuned for unconstrained video data (as opposed to a few captive individuals), making it especially suitable for the raw field footage used in conservation, ecology, and behavioural research.
\vspace{-2ex}

\subsection{Contrastive SimCLR versus DINOv2}
\vspace{-1ex}
\label{sec:contrastive_comparison}
To benchmark a different self-supervised strategy, we trained a SimCLR ViT-S on the same \textit{L+P+B} clips, using a batch size of 1024 faces and applying gradient checkpointing to 80\,\% of the model.  
SimCLR achieves 62.1\,\% Re-ID on Bossou-9 and 16.0\,\% on PetFaceC$\star$, lagging behind DINOv2 by 16.0 and 23.3 pp, respectively.  
These results confirm that the DINOv2 approach is substantially more effective than plain contrastive learning for challenging, in-the-wild chimpanzee data.
\vspace{-1ex}

\subsection{Ablation: Training Mix}
\vspace{-1ex}
\label{sec:ablation_mix_revised}
Adding the diverse PanAf footage (P) to Loma data (L) boosts Bossou-9 Re-ID from 69.4\,\% to 74.6\,\%.  
Introducing the \emph{unlabelled} Bossou-14 clips (B) yields a further +3.5 pp on Bossou-9 and +3.6 pp on PetFaceC$\star$, illustrating how self-supervision readily exploits additional video, even without identity annotations, and generalises to unseen chimpanzees.
We observed that incorporating the lower-quality PanAf (P) data results in a modest drop in performance on the higher-quality PetFaceC$\star$ benchmark, likely reflecting a distribution mismatch in image quality between training and evaluation.
\vspace{-1ex}

\subsection{Model Scaling and Longer Pre‑training}
\vspace{-0.5ex}
\label{sec:model_scaling}
A larger ViT-B backbone (87 M parameters) and double the optimisation steps add +4.1 pp on Bossou-9 Re-ID and push verification to 76.3\,\%.  
These gains indicate headroom for still larger models and longer schedules, especially as ever-growing wildlife video archives become available.

\section{Summary and Extensions}
\label{sec:summary_extensions}

Our work has demonstrated that a fully self‑supervised Vision Transformer trained with the DINOv2 objective on automatically mined chimpanzee face tracks can act as a universal, open‑set face embedder that outperforms fully supervised baselines on a challenging re‑identification benchmark, Bossou, and approaches their performance on in-captivity chimps in PetFace, while requiring \emph{no} manual identity labels.

Metadata accompanying each camera‑trap sequence, e.g.\ recording time, geographic location and the track continuity returned by the detector-tracker pipeline, could further provide reliable positive or negative pairs. 
Such auxiliary signals would allow us to mine relational constraints (two face crops that appear in the same short track are almost certainly the \emph{same} individual; faces taken at overlapping times from distant cameras are likely \emph{different}) without additional human annotation. 
Although our present method does not exploit these cues, integrating them into the loss, e.g.\ through weighted neighbourhood sampling or curriculum mining, offers a promising avenue for future work.

Another extension is to combine the self-distillation self-supervised losses with an explicit \textit{self‑supervised contrastive} term, e.g.\ \cite{zhang2022c3,wolf2023self,hernandez2024vic}.  Adapting such training recipes to camera‑trap footage, possibly with metadata‑guided positive/negative mining, could further strengthen open‑set individual face recognition.

Overall, utilising our self-supervised approach and automatic mining of chimpanzee face tracks creates a pathway toward open-set face recognition in ecological contexts — mirroring capabilities long available for human datasets. Such tools have the potential to significantly reduce reliance on costly manual annotation and enable scalable, low-cost monitoring of wild ape populations, supporting long-term behavioural, demographic, and conservation research.

\vfill\pagebreak
\label{sec:refs}

\small
\bibliographystyle{IEEEbib}
\bibliography{refs}

\begin{thebibliography}{10}

\bibitem{debetencourt2024camera}
B.~Debetencourt, M.~M. Barry, M.~Arandjelovic, C.~Stephens, N.~Maldonado, and C.~Boesch,
\newblock ``Camera traps unveil demography, social structure, and home range of six unhabituated western chimpanzee groups in the moyen bafing national park, guinea,''
\newblock {\em American Journal of Primatology}, 2024.

\bibitem{guo2020automatic}
S.~Guo, P.~Xu, Q.~Miao, G.~Shao, C.~A. Chapman, X.~Chen, G.~He, D.~Fang, H.~Zhang, Y.~Sun, et~al.,
\newblock ``Automatic identification of individual primates with deep learning techniques,''
\newblock {\em iScience}, 2020.

\bibitem{wang2018cosface}
H.~Wang, Y.~Wang, Z.~Zhou, X.~Ji, D.~Gong, J.~Zhou, Z.~Li, and W.~Liu,
\newblock ``{CosFace}: Large margin cosine loss for deep face recognition,''
\newblock in {\em CVPR}, 2018.

\bibitem{deng2019arcface}
J.~Deng, J.~Guo, N.~Xue, and S.~Zafeiriou,
\newblock ``{ArcFace}: Additive angular margin loss for deep face recognition,''
\newblock in {\em CVPR}, 2019.

\bibitem{guo2016ms}
Y.~Guo, L.~Zhang, Y.~Hu, X.~He, and J.~Gao,
\newblock ``{MS-Celeb-1M}: A dataset and benchmark for large-scale face recognition,''
\newblock in {\em ECCV}, 2016.

\bibitem{freytag2016chimpanzee}
A.~Freytag, E.~Rodner, M.~Simon, A.~Loos, H.~S. K{\"u}hl, and J.~Denzler,
\newblock ``Chimpanzee faces in the wild: Log-euclidean {CNNs} for predicting identities and attributes of primates,''
\newblock in {\em GCPR}, 2016.

\bibitem{grother2003face}
P.~Grother, R.~Micheals, and P.~Phillips,
\newblock ``Face recognition vendor test 2002 performance metrics,''
\newblock in {\em International conference on audio-and video-based biometric person authentication}, 2003.

\bibitem{scheirer2012toward}
Walter~J Scheirer, Anderson de~Rezende~Rocha, Archana Sapkota, and Terrance~E Boult,
\newblock ``Toward open set recognition,''
\newblock {\em IEEE TPAMI}, 2012.

\bibitem{caron2021emerging}
M.~Caron, H.~Touvron, I.~Misra, H.~J{\'e}gou, J.~Mairal, P.~Bojanowski, and A.~Joulin,
\newblock ``Emerging properties in self-supervised vision transformers,''
\newblock in {\em ICCV}, 2021.

\bibitem{oquab2023dinov2}
M.~Oquab, T.~Darcet, T.~Moutakanni, H.~Vo, M.~Szafraniec, V.~Khalidov, P.~Fernandez, D.~Haziza, F.~Massa, A.~El-Nouby, et~al.,
\newblock ``{DINOv2}: Learning robust visual features without supervision,''
\newblock {\em arXiv}, 2023.

\bibitem{loos2013automated}
A.~Loos and A.~Ernst,
\newblock ``An automated chimpanzee identification system using face detection and recognition,''
\newblock {\em EURASIP Journal on Image and Video Processing}, 2013.

\bibitem{deb2018face}
D.~Deb, S.~Wiper, S.~Gong, Y.~Shi, C.~Tymoszek, A.~Fletcher, and A.~K. Jain,
\newblock ``Face recognition: Primates in the wild,''
\newblock in {\em BTAS}, 2018.

\bibitem{schofield2019chimpanzee}
D.~Schofield, A.~Nagrani, A.~Zisserman, M.~Hayashi, T.~Matsuzawa, D.~Biro, and S.~Carvalho,
\newblock ``Chimpanzee face recognition from videos in the wild using deep learning,''
\newblock {\em Science Advances}, 2019.

\bibitem{he2016deep}
K.~He, X.~Zhang, S.~Ren, and J.~Sun,
\newblock ``Deep residual learning for image recognition,''
\newblock in {\em CVPR}, 2016.

\bibitem{shinoda2024petface}
R.~Shinoda and K.~Shiohara,
\newblock ``Petface: A large-scale dataset and benchmark for animal identification,''
\newblock in {\em ECCV}, 2024.

\bibitem{liu2021swin}
Z.~Liu, Y.~Lin, Y.~Cao, H.~Hu, Y.~Wei, Z.~Zhang, S.~Lin, and B.~Guo,
\newblock ``Swin transformer: Hierarchical vision transformer using shifted windows,''
\newblock in {\em ICCV}, 2021.

\bibitem{vcermak2024wildlifedatasets}
V.~{\v{C}}erm{\'a}k, L.~Picek, L.~Adam, and K.~Papafitsoros,
\newblock ``Wildlifedatasets: An open-source toolkit for animal re-identification,''
\newblock in {\em WACV}, 2024.

\bibitem{WildMe2023}
L.~Otarashvili,
\newblock ``Miewid,'' 2023.

\bibitem{Laskowski2023GorillaVision}
L.~Laskowski, R.~Sawahn, M.~Schall, D.~Wasmuht, M.~Bermejo, and G.~de~Melo,
\newblock ``Gorillavision -- open-set re-identification of wild gorillas,''
\newblock in {\em CamTrap WS '23 Workshop}, 2023.

\bibitem{douze2024faiss}
M.~Douze, A.~Guzhva, C.~Deng, J.~Johnson, G.~Szilvasy, P.-E. Mazar{\'e}, M.~Lomeli, L.~Hosseini, and H.~J{\'e}gou,
\newblock ``The {FAISS} library,''
\newblock {\em arXiv}, 2024.

\bibitem{chen2020simple}
T.~Chen, S.~Kornblith, M.~Norouzi, and G.~Hinton,
\newblock ``A simple framework for contrastive learning of visual representations,''
\newblock in {\em ICML}, 2020.

\bibitem{chen2021empirical}
X.~Chen, S.~Xie, and K.~He,
\newblock ``An empirical study of training self-supervised vision transformers,''
\newblock in {\em ICCV}, 2021.

\bibitem{he2022masked}
K.~He, X.~Chen, S.~Xie, Y.~Li, P.~Doll{\'a}r, and R.~Girshick,
\newblock ``Masked autoencoders are scalable vision learners,''
\newblock in {\em CVPR}, 2022.

\bibitem{zhou2021ibot}
J.~Zhou, C.~Wei, H.~Wang, W.~Shen, C.~Xie, A.~Yuille, and T.~Kong,
\newblock ``{iBOT}: Image {BERT} pre-training with online tokenizer,''
\newblock {\em ICLR}, 2022.

\bibitem{kuzdeuov2023anyface}
A.~Kuzdeuov, D.~Koishigarina, and H.~A. Varol,
\newblock ``{AnyFace}: A data-centric approach for input-agnostic face detection,''
\newblock in {\em BigComp}, 2023.

\bibitem{ge2021yolox}
Z.~Ge, S.~Liu, F.~Wang, Z.~Li, and J.~Sun,
\newblock ``{YOLOX}: Exceeding {YOLO} series in 2021,''
\newblock {\em arXiv}, 2021.

\bibitem{zhang2022bytetrack}
Y.~Zhang, P.~Sun, Y.~Jiang, D.~Yu, F.~Weng, Z.~Yuan, P.~Luo, W.~Liu, and X.~Wang,
\newblock ``{ByteTrack}: Multi-object tracking by associating every detection box,''
\newblock in {\em ECCV}, 2022.

\bibitem{brookes2024panaf20k}
O.~Brookes, M.~Mirmehdi, C.~Stephens, S.~Angedakin, K.~Corogenes, D.~Dowd, P.~Dieguez, T.~C. Hicks, S.~Jones, K.~Lee, et~al.,
\newblock ``{PanAf20k}: A large video dataset for wild ape detection and behaviour recognition,''
\newblock {\em IJCV}, 2024.

\bibitem{molina2023reaffirming}
G.~Molina-Vacas, R.~Mu{\~n}oz-Mas, B.~Amarasekaran, and R.~M. Garriga,
\newblock ``Reaffirming the loma mountains national park in sierra leone as a critical site for the conservation of west african chimpanzee (pan troglodytes verus),''
\newblock {\em American Journal of Primatology}, 2023.

\bibitem{zhang2022c3}
C.~Zhang and D.~Yu,
\newblock ``{C3-DINO}: Joint contrastive and non-contrastive self-supervised learning for speaker verification,''
\newblock {\em IEEE Journal of Selected Topics in Signal Processing}, 2022.

\bibitem{wolf2023self}
D.~Wolf, T.~Payer, C.~S. Lisson, C.~G. Lisson, M.~Beer, M.~G{\"o}tz, and T.~Ropinski,
\newblock ``Self-supervised pre-training with contrastive and masked autoencoder methods for dealing with small datasets in deep learning for medical imaging,''
\newblock {\em Scientific Reports}, 2023.

\bibitem{hernandez2024vic}
J.~Hernandez, R.~Villegas, and V.~Ordonez,
\newblock ``{Vic-MAE}: Self-supervised representation learning from images and video with contrastive masked autoencoders,''
\newblock in {\em ECCV}, 2024.

\end{thebibliography}

\end{document}